\documentclass{llncs}
\usepackage{graphicx}
\usepackage[linesnumbered]{algorithm2e}

\title{Generating Difficult SAT Instances by Preventing Triangles}
\author{Guillaume Escamocher \and Barry O'Sullivan \and Steven David Prestwich}
\institute{INSIGHT Centre for Data Analytics\\University College Cork\\Cork, Ireland}

\begin{document}

\maketitle

\begin{abstract}

When creating benchmarks for SAT solvers, we need SAT instances that
are easy to build but hard to solve. A recent development in the
search for such methods has led to the Balanced SAT algorithm, which
can create $k$-SAT instances with $m$ clauses of high difficulty, for
arbitrary $k$ and $m$. In this paper we introduce the No-Triangle SAT
algorithm, a SAT instance generator based on the cluster coefficient
graph statistic. We empirically compare the two algorithms by fixing
the arity and the number of variables, but varying the number of
clauses. The hardest instances that we find are produced by
No-Triangle SAT. Furthermore, difficult instances from No-Triangle SAT
have a different number of clauses than difficult instances from
Balanced SAT, potentially allowing a combination of the two methods to
find hard SAT instances for a larger array of parameters.

\end{abstract}

\section{Introduction}

The Boolean Satisfiability Problem, commonly called SAT, is the
problem of assigning values to a set of variables while satisfying a
set of disjunctive clauses, each consisting of literals representing
either a variable or its negation. State-of-the-art SAT solvers are
commonly tested on benchmarks of various sizes and
complexities. Creating these benchmarks requires the construction of
many instances that present a challenge to the solvers. Ideally this
challenge should not come exclusively from a prohibitively high total
number of literals. Indeed, at least some instances of moderate and
even small sizes should be considered hard to solve. For this reason,
and more generally to help advance the understanding of instance
complexity in constraint problems, it is important to have a way to
create at will difficult instances of any size.

In major SAT competitions~\cite{satcompetition}, the smallest
instances that cannot be solved within the time limit are obtained by
hiding cardinality constraints. They offer a great deal of difficulty
to solvers when presented for the first time, but are usually
exploitable by a particular trick when identified. As an illustration,
the sgen6 algorithm is able to generate instances with less than a
thousand literals that no solver can solve in less than 10
minutes~\cite{sgen}, but these instances can be reduced to a simple
matching problem~\cite{knuthtrick}.

A more general drawback of this kind of generators is that they are
not easily parameterizable. Because of their rigid structure, fixing
one parameter (such as the number of variables) constrains the other
parameters (such as the arity or the number of clauses) to very
specific values. The sgen6 algorithm, for example, outputs instances
with clauses of different arity, including clauses with only two
literals, and with a fairly low number of clauses compared to the
number of variables (about twice as many).

To address these issues, a new method of generating difficult
instances has been proposed, called Balanced
SAT~\cite{balancedsat}. The main idea of Balanced SAT is to balance
the number of occurrences of each literal, as well as minimizing the
number of variable pairs that appear in different clauses. Instances
created by Balanced SAT are not as difficult to solve as the hardest
crafted instances, but because of their random nature they are not as
easily exploitable. Furthermore, Balanced SAT can generate instances
of any arity, any number of variables and any number of clauses.

Constraint problems exhibit a phase transition phenomenon, in which
instances under a given constrainedness threshold are increasingly
likely to be satisfiable as the size increases, while instances with
greater constrainedness are increasingly likely to be
unsatisfiable~\cite{phasetransition}. Empirically, instances close to
this threshold are found to be extremely hard, while under- and
over-constrained instances are found to be much easier.

For SAT instances with exactly 3 literals in each clause, the
threshold is conjectured to be approximately
$m=4.267n$~\cite{threshold}, where $n$ is the number of variables and
$m$ is the number of clauses. For completely random instances this
ratio is where the hard instances are. The peak of difficulty for
instances built by generators of hard instances is distinctly
lower. As mentioned above, the ratio for sgen6 instances is about 2,
while for Balanced SAT it is approximately
3.6~\cite{balancedsat}. These low figures might possibly be explained
by the fact that instance generators of this kind try to force solvers
to look at a large number of possible variable assignments, and adding
more clauses decreases the number of paths to explore. In any case, we
are not aware of any existing generator of difficult instances with an
observed peak in the over-constrained region of random instances.

Our main contribution in this paper is a new algorithm for generating
SAT instances, called No-Triangle SAT. Like Balanced SAT, No-Triangle
SAT can generate instances of any arity, any number of variables and
any number of clauses. It balances the number of occurrences of each
literal, again like Balanced SAT, but in addition of only avoiding
redundant variable pairs, it also minimizes the number of constraint
triangles in the constraint graph associated to the
instance. Constraint triangles, defined in
Section~\ref{sec:defpreliminary}, are related to the cluster
coefficient measure of the constraint graph.

While several parts of our No-Triangle algorithm are directly inspired
from Balanced SAT, we show in Section~\ref{sec:empire} that the
behavior of instances obtained from these two generators is
drastically different in two aspects. First, the peak of difficulty
for No-Triangle SAT instances is taller than the one for Balanced SAT
instances. Additionally, for a fixed arity and a fixed number of
variables, difficult No-Triangle SAT instances appear at a much higher
number of clauses than Balanced SAT instances. No-Triangle SAT thus
constitutes a way to build highly constrained instances that are hard
to solve.

In the next section we describe both the existing Balanced SAT
generator and our novel No-Triangle SAT generator. We also define some
graph notions that are integral to our algorithm. In
Section~\ref{sec:empire} we present the core results of the paper,
empirical studies for different instance sizes of the behavior of
No-Triangle SAT instances compared to Balanced SAT as well as random
instances. Finally we conclude in Section~\ref{sec:conclusion}.

\section{Generators}\label{sec:definitions}

\subsection{Preliminary notions}\label{sec:defpreliminary}

We begin by formally defining the Boolean Satisfiability Problem
(SAT).

\begin{definition}\label{def:satinstance}
A \emph{SAT instance} is composed of $n$ variables $v_1,v_2,\dots,v_n$
and $m$ clauses $C_1,C_2,\dots,C_m$, where each clause $C_i$ is a
disjunction of $k_i$ literals $l_1\vee l_2\vee\dots\vee l_{k_i}$ and
each literal is either a variable from $\{v_1,v_2,\dots,v_n\}$ or the
negation of one such variable. The \emph{arity} of the SAT instance is
the number of literals in the clause with the most literals.
\end{definition}

A literal is \emph{positive} if it corresponds to a variable, and
\emph{negative} if it corresponds to the negation of a variable. The
\emph{polarity} of a literal is its sign. A \emph{$k$-SAT instance} is
a SAT instance with exactly $k$ literals in each clause. A
\emph{solution} for a ($k$-)SAT instance $I$ is an assignment of
boolean values to all $n$ variables such that all $m$ clauses of $I$
are satisfied.

The problem of determining whether a given SAT instance admits a
solution was the first to be shown NP-Complete~\cite{sat}. Binary SAT
instances are polynomial~\cite{2sat} but allowing even just 3 literals
in each clause (3-SAT) is known to be NP-Complete~\cite{3sat}. Since
its inception in Karp's 21 NP-Complete problems, 3-SAT has in fact
been a popular problem to reduce from in NP-hardness proofs. Most
of our paper is therefore focused on 3-SAT instances, however both the
Balanced SAT algorithm and our own No-Triangle generator can be used
to generate SAT instances of any arity.

The intuition behind our method is to minimize the amount of
similarities between clauses. We present a few concepts to help
structure this idea. We start by the notion of repeated pairs of
variables, which is also used by Balanced SAT.

\begin{definition}\label{def:repeatedpair}
Let $I$ be a SAT instance. Let $v$ and $v'$ be two variables of
$I$. We say that $v$ and $v'$ form a \emph{repeated pair} if they
occur together in at least two different clauses of $I$, regardless of
the polarity of their literals.
\end{definition}

To benefit from several Graph Theory properties, we view a SAT
instance as a graph.

\begin{definition}\label{def:constraintgraph}
Let $I$ be a SAT instance. The \emph{constraint graph} of $I$ is the
graph $G$ such that the vertices of $G$ are the variables of $I$ and
the edges of $G$ are the pairs of variables of $I$ that occur together
in a same clause, regardless of the polarity of their literals.
\end{definition}

While a repeated pair of variables forms an edge in the constraint
graph, not all constraint graph edges are repeated pairs.

Our No-Triangle SAT algorithm does not just study pairwise
relations. It goes one step further and also looks at three-sided
transitive structures.

\begin{definition}\label{def:constrainttriangle}
Let $I$ be a SAT instance and let $v_1$, $v_2$ and $v_3$ be three
variables of $I$. We say that $v_1$, $v_2$ and $v_3$ form a
\emph{constraint triangle} if the three pairs $\langle
v_1,v_2\rangle$, $\langle v_1,v_3\rangle$ and $\langle v_2,v_3\rangle$
are edges in the constraint graph of $I$. If exactly two out of these
three pairs are edges in the constraint graph of $I$, we instead say
that $v_1$, $v_2$ and $v_3$ form an \emph{incomplete constraint
  triangle}.
\end{definition}

Note that it is possible for three variables to form a constraint
triangle even if no single clause contains all three of them. The
notion of constraint triangle is related to the cluster coefficient
graph metric, one of the measures characterizing small-world
networks~\cite{smallworld}.


\subsection{Balanced SAT}

When $k$, $n$ and $m$ are the desired arity, number of variables and
number of clauses respectively, the Balanced SAT
algorithm~\cite{balancedsat} creates $\frac{k\times m}{n}$ rows of
variables, where each row contains every variable exactly once (except
the last row if $k\times m$ is not a multiple of $n$). The variables
are then sorted within each row by a greedy heuristic that picks the
variable that minimizes the number of variable pairs in the current
clause that are already edges in the constraint graph. Finally, the
polarities are assigned randomly for the first occurrence of each
variable, and alternatively for the remaining occurrences.

A pseudocode representation of Balanced SAT is given by
Algorithm~\ref{alg:balanced}.

\begin{algorithm}
\caption{\label{alg:balanced}Balanced SAT generator.}
\KwData{Three integers $k$, $n$ and $m$.}
\KwResult{A $k$-SAT instance with $n$ variables $v_1,v_2,\dots,v_n$ and $m$ clauses $C_1,C_2,\dots,C_m$.}
Start with $I$, a SAT instance with $m$ empty clauses $C_1,C_2,\dots,C_m$\;
\For{$i\leftarrow 1$ \KwTo $m$}
    {\For{$j\leftarrow 1$ \KwTo $k$}
        {Pick the variable $v$ that occurs the fewest in $I$ so far\;
        Break ties by minimizing the amount of repeated pairs added by introducing $v$ in $C_i$\;
        Add the literal $v$ to $C_i$\;
        Increment the number of occurrences of $v$ by 1\;
        Update the database of variable pairs present in $I$\;}}
\For{$i\leftarrow 1$ \KwTo $n$}
    {With probability $\frac{1}{2}$, change the first occurrence of $v_i$ in $I$ to a negative literal\;
    Set each subsequent occurrence of $v_i$ to the negation of the previous occurrence\;}
\Return $I$\;
\end{algorithm}

\subsection{No-Triangle SAT}

Considering only repeated pairs for discriminating between variables
can still leave several potential candidates, and Balanced SAT has no
choice but to pick one of them randomly. No-Triangle SAT introduces an
additional tie-breaker: the number of constraint triangles formed by
adding the variable considered to the current clause. This corresponds
to Lines 6 and 10 in Algorithm~\ref{alg:notriangle}. While seemingly
only a minor alteration, we shall show in the next section that
No-Triangle SAT instances behave very differently than Balanced SAT
ones.

\begin{algorithm}
\caption{\label{alg:notriangle}No-Triangle SAT generator.}
\KwData{Three integers $k$, $n$ and $m$.}
\KwResult{A $k$-SAT instance with $n$ variables $v_1,v_2,\dots,v_n$ and $m$ clauses $C_1,C_2,\dots,C_m$.}
Start with $I$, a SAT instance with $m$ empty clauses $C_1,C_2,\dots,C_m$\;
\For{$i\leftarrow 1$ \KwTo $m$}
    {\For{$j\leftarrow 1$ \KwTo $k$}
        {Pick the variable $v$ that occurs the fewest in $I$ so far\;
        Break ties by minimizing the amount of repeated pairs added by introducing $v$ in $C_i$\;
        Break further ties by minimizing the amount of constraint triangles added by introducing $v$ in $C_i$\;
        Add the literal $v$ to $C_i$\;
        Increment the number of occurrences of $v$ by 1\;
        Update the database of variable pairs present in $I$\;
        Update the database of incomplete constraint triangles present in $I$\;}}
\For{$i\leftarrow 1$ \KwTo $n$}
    {With probability $\frac{1}{2}$, change the first occurrence of $v_i$ in $I$ to a negative literal\;
    Set each subsequent occurrence of $v_i$ to the negation of the previous occurrence\;}
\Return $I$\;
\end{algorithm}

When trying to generate hard instances, we avoid constraint triangles
and encourage incomplete constraint triangles. This mirrors results in
Constraint Satisfaction Problems, where it has been shown that the
absence of patterns similar to incomplete constraint triangles
fulfills the Joint-Winner Property and constitutes a tractable class,
while merely forbidding complete constraint triangles still yields an
NP-Complete complexity class~\cite{jointwinner}.

\section{Experimental results}\label{sec:empire}

\begin{figure}
\centering
\begin{minipage}{.49\textwidth}
\centering
\includegraphics[scale=.35]{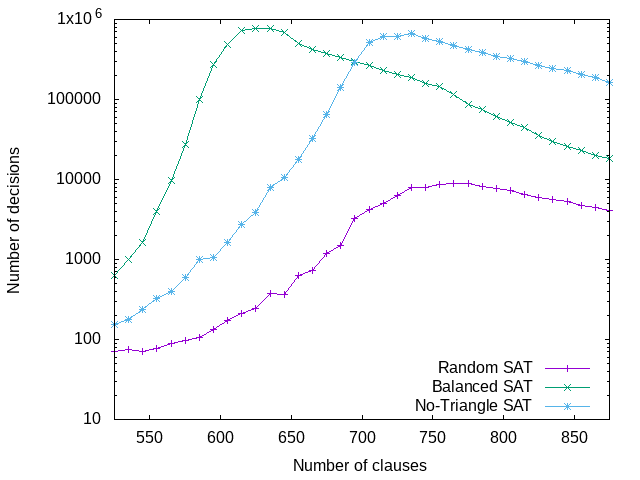}
\caption{Instance difficulty for 3-SAT instances with 175 variables.}
\label{fig:peaks175}
\end{minipage}
\begin{minipage}{.49\textwidth}
\centering
\includegraphics[scale=.35]{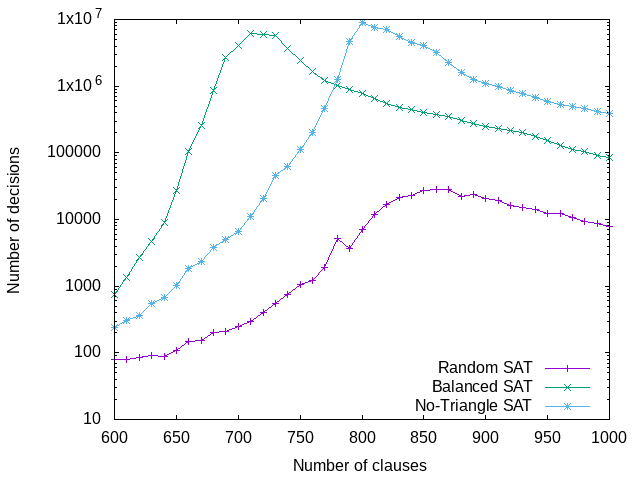}
\caption{Instance difficulty for 3-SAT instances with 200 variables.}
\label{fig:peaks200}
\end{minipage}
\end{figure}
\begin{figure}
\centering
\includegraphics[scale=.35]{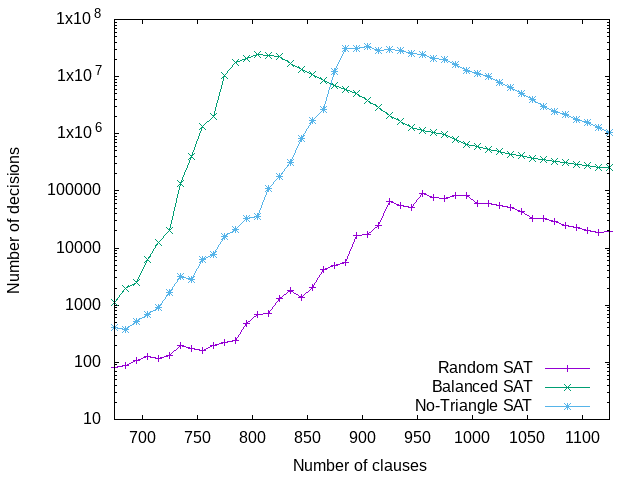}
\caption{Instance difficulty for 3-SAT instances with 225 variables.}
\label{fig:peaks225}
\end{figure}

We empirically compared three different ways to generate 3-SAT
instances with $n$ variables and $m$ clauses. The first is Random SAT,
where every literal in each clause is randomly picked from the $2n$
possible choices, with no influence from previous picks, with the
exception of not allowing the exact same clause twice. The second
method is Balanced SAT, and the last one is our No-Triangle SAT. We
tested the three algorithms on sizes $n=175$, $n=200$ and $n=225$,
with the results presented in Figure~\ref{fig:peaks175},
\ref{fig:peaks200} and \ref{fig:peaks225} respectively. To capture the
most interesting instances, the ones around the peak of difficulty, we
varied the number of clauses from $m=3n$ to $m=5n$, with a step of
10. The X axis represents the number of clauses, while the Y axis
represents the number of decisions made by the solver. Each point in
Figures~\ref{fig:peaks175} and~\ref{fig:peaks225} represents the
average of 100 instances, while each point in
Figure~\ref{fig:peaks200} represents the average of 25 instances. The
solver we used was the 2018 SAT competition version of
CaDiCaL~\cite{cadical}.

The general behavior is the same for all three sizes: the peak of
difficulty for No-Triangle SAT is taller than the one for Balanced SAT
(with the exception of $n=175$) and occurs at a much higher number of
clauses. In fact, for instances with at least four times as many
clauses as variables, No-Triangle SAT instances are for all three
sizes one order of magnitude harder then Balanced SAT instances.

We suspect that as the number of variables increases, No-Triangle SAT
performs better in comparison to Balanced SAT. Indeed, the maximum
possible number of edges in a graph is quadratic in the number of
vertices, but because the arity $k$ of the instances is fixed, the
number of actually occurring variable pairs is linear (equal to $3m$
for $k=3$). Therefore it is easier to avoid repeated pairs for larger
values of $n$, and the additional tie-breaker from No-Triangle SAT
will be used more often. This would explain why in our experiments the
peak of difficulty for No-Triangle SAT seems to grow faster as $n$
increases than the one for Balanced SAT.

Of course, looking only at average results does not always give the
full picture. Even when fixing both the number of variables and the
number of clauses, hardness between instances can vary
considerably. Near the peak of difficulty has been observed the
presence of a few instances that are extremely hard for backtrack
procedures, as well as many instances that are much easier. This
phenomenon is called heavy-tailed behavior~\cite{heavytail}. For this
reason, it is important when comparing difficult instance generators
to also consider the instances that diverge the most from the norm.

We give more detailed numbers in Table~\ref{tab:extremes}\footnote{Full results are available in Appendix~\ref{sec:appendix}.}. Of
particular interest are the lines labelled ``Easiest'', which indicate
the hardest instance obtained when only keeping the instance with the
fewest number of decisions for each number of clauses. As an example,
the easiest instance generated by No-Triangle SAT with parameters
$n=200$ and $m=850$ required 2,161,557 decisions to be solved, and for
every other number of clauses tested between 600 and 1,000 there was
at least one instance with the same number of variables generated by
No-Triangle SAT that was easier.


\begin{table}[t]
\centering
\caption{Comparison of the extremums reached by the generators.}
\begin{tabular}{|c|c||r|r|r|r|r|r|}
\hline
\multicolumn{2}{|c||}{} & \multicolumn{2}{c|}{Random SAT} & \multicolumn{2}{c|}{Balanced SAT} & \multicolumn{2}{c|}{No-Triangle SAT}\\
\cline{3-8}
\multicolumn{2}{|c||}{} & \#clauses & \#decisions & \#clauses & \#decisions & \#clauses & \#decisions\\
\hline
\hline
 & Easiest & 805 & 3,138 & 645 & $\mathbf{407,930}$ & 775 & 341,182\\
$n=175$ & Average & 765 & 9,021 & 635 & $\mathbf{782,832}$ & 735 & 666,457\\
& Hardest & 715 & 46,144 & 595 & 2,944,991 & 685 & $\mathbf{2,984,858}$\\
\hline
 & Easiest & 910 & 6,457 & 760 & 868,907 & 850 & $\mathbf{2,161,557}$\\
$n=200$ & Average & 870 & 28,202 & 710 & 6,259,566 & 800 & $\mathbf{8,880,628}$\\
& Hardest & 840 & 125,413 & 690 & 17,891,392 & 790 & $\mathbf{22,012,457}$\\
\hline
 & Easiest & 1075 & 28,137 & 835 & 10,057,838 & 935 & $\mathbf{24,184,043}$\\
$n=225$ & Average & 955 & 90,843 & 805 & 24,711,737 & 905 & $\mathbf{34,320,235}$\\
 & Hardest & 955 & 245,304 & 775 & $\mathbf{65,268,387}$ & 875 & 63,315,167\\
\hline
\end{tabular}
\label{tab:extremes}
\end{table}

The behavior of the easiest instance at each number of clauses is very
meaningful, because it is key to being able to reliably generate hard
instances without having to solve them. We know for instance that
Balanced SAT can generate instances with $n=225$ and $m=775$ that
require more than 60 million decisions to solve. However, during the
course of our experiments Balanced SAT also generated an instance with
the same parameters that only required 148,000 decisions to
solve. Similarly, while we can expect No-Triangle SAT to provide for
$n=225$ and $m=905$ instances with an average difficulty of 34 million
decisions, we also obtained an instance with these same parameters
that required about 1.3 million decisions to solve. Using the
algorithms to get a truly difficult instance with these values for $n$
and $m$ is possible, but it would require solving the instances
generated to ensure that they are hard enough. And because of the
heavy-tailed behavior, creating a large set of instances would not
guarantee that even half of them fulfill the difficulty
requirements. In contrast, the easiest instance for some fixed
parameters gives a measure of guaranteed difficulty.


When only keeping the easiest instance at each number of clauses, the
difference between the two generators is stark. For $n=200$ and
$n=225$, the hardest such No-Triangle SAT instance requires more than
twice as many decisions as its Balanced SAT counterpart. In fact, the
\emph{easiest} No-Triangle SAT instance with parameters $n=225$ and
$m=935$ is about as difficult to solve as the \emph{average} instance
with the same number of variables found at the peak of average
difficulty for Balanced SAT. This shows the great usefulness of
No-Triangle SAT when needing to quickly generate many hard instances
for a given number of variables.


\section{Conclusion}\label{sec:conclusion}

We have introduced No-Triangle SAT, an algorithm that can generate SAT
instances with any arity, number of variables and/or number of
clauses. Compared to existing generators with the same properties,
like Balanced SAT, No-Triangle SAT can create instances that are
harder, especially when asked to provide heavily constrained
instances. Moreover, when fixing the number of variables No-Triangle
SAT can build without needing to solve them instances with a high
guaranteed difficulty.

We believe that future work on constraint triangles could help solver
heuristics for instances with a large number of clauses. Since
Balanced SAT still performs well for instances with fewer clauses, one
could also imagine combining the two algorithms, using Balanced SAT
for underconstrained instances and No-Triangle SAT for instances with
more clauses.

\bibliographystyle{plain}
\bibliography{SAT_Generation}

\newpage
\appendix
\section{Complete results}\label{sec:appendix}

In all nine following tables, each column represents the following data:

\begin{itemize}
\item[Column  1:]  Number of clauses
\item[Column  2:]  Number of runs
\item[Column  3:]  Number of satisfiable instances
\item[Column  4:]  Number of decisions, minimum
\item[Column  5:]  Number of decisions, total (divide by Column 2 to get the average)
\item[Column  6:]  Number of decisions, maximum
\item[Column  7:]  Number of repeated pairs, minimum
\item[Column  8:]  Number of repeated pairs, total (divide by Column 2 to get the average)
\item[Column  9:]  Number of repeated pairs, maximum
\item[Column 10:]  Average distance between variables, minimum
\item[Column 11:]  Average distance between variables, total (divide by Column 2 to get the average)
\item[Column 12:]  Average distance between variables, maximum
\item[Column 13:]  Cluster coefficient, minimum
\item[Column 14:]  Cluster coefficient, total (divide by Column 2 to get the average)
\item[Column 15:]  Cluster coefficient, maximum
\end{itemize}

\newpage

\begin{table}
\centering
\caption{Full results of the Random SAT algorithm for N=175.}
\begin{tabular}{|r|r|r|r|r|r|r|r|r|r|r|r|r|r|r|}
\hline
525 & 100 & 100 & 45 & 7067 & 149 & 62 & 7784 & 107 & 2.08 & 21109 & 4.05 & 0.143 & 15005 & 0.157\\
535 & 100 & 100 & 43 & 7504 & 220 & 61 & 8132 & 106 & 2.06 & 21387 & 4.06 & 0.144 & 15107 & 0.158\\
545 & 100 & 100 & 37 & 7035 & 154 & 64 & 8272 & 103 & 2.05 & 20862 & 4.03 & 0.144 & 15183 & 0.160\\
555 & 100 & 100 & 43 & 7765 & 147 & 63 & 8745 & 108 & 2.04 & 20745 & 4.01 & 0.144 & 15255 & 0.160\\
565 & 100 & 100 & 45 & 9001 & 247 & 70 & 9104 & 109 & 2.03 & 20646 & 4.01 & 0.148 & 15362 & 0.163\\
575 & 100 & 100 & 42 & 9633 & 428 & 69 & 9408 & 113 & 2.02 & 20531 & 4.01 & 0.144 & 15431 & 0.161\\
585 & 100 & 100 & 42 & 10635 & 419 & 80 & 9796 & 124 & 2.01 & 20223 & 2.04 & 0.148 & 15498 & 0.161\\
595 & 100 & 100 & 39 & 13185 & 849 & 82 & 10206 & 124 & 2.00 & 20346 & 3.98 & 0.148 & 15657 & 0.165\\
605 & 100 & 100 & 42 & 17174 & 772 & 77 & 10160 & 127 & 1.99 & 20034 & 2.02 & 0.151 & 15733 & 0.167\\
615 & 100 & 100 & 51 & 21164 & 854 & 85 & 10553 & 137 & 1.98 & 19950 & 2.01 & 0.149 & 15868 & 0.166\\
625 & 100 & 100 & 38 & 24527 & 1406 & 90 & 11056 & 130 & 1.97 & 20057 & 3.96 & 0.152 & 15928 & 0.166\\
635 & 100 & 100 & 46 & 37855 & 4172 & 91 & 11577 & 139 & 1.96 & 19789 & 1.99 & 0.152 & 16071 & 0.168\\
645 & 100 & 100 & 52 & 36827 & 2513 & 94 & 11745 & 137 & 1.96 & 19706 & 1.98 & 0.154 & 16157 & 0.169\\
655 & 100 & 100 & 39 & 62257 & 3246 & 93 & 12213 & 156 & 1.95 & 19621 & 1.98 & 0.156 & 16320 & 0.169\\
665 & 100 & 100 & 53 & 73574 & 6262 & 107 & 12776 & 158 & 1.94 & 19943 & 3.94 & 0.157 & 16395 & 0.173\\
675 & 100 & 100 & 69 & 120159 & 8750 & 102 & 12855 & 153 & 1.94 & 19684 & 3.92 & 0.156 & 16515 & 0.172\\
685 & 100 & 100 & 44 & 150058 & 13408 & 103 & 13189 & 161 & 1.93 & 19402 & 1.96 & 0.160 & 16579 & 0.173\\
695 & 100 & 99 & 36 & 325619 & 27266 & 103 & 13461 & 161 & 1.92 & 19335 & 1.95 & 0.158 & 16697 & 0.174\\
705 & 100 & 95 & 45 & 420422 & 32231 & 114 & 13924 & 160 & 1.92 & 19284 & 1.94 & 0.163 & 16830 & 0.174\\
715 & 100 & 90 & 111 & 497117 & 46144 & 119 & 14352 & 170 & 1.91 & 19214 & 1.93 & 0.162 & 16946 & 0.176\\
725 & 100 & 81 & 41 & 623711 & 32183 & 124 & 14765 & 174 & 1.91 & 19172 & 1.93 & 0.164 & 17092 & 0.180\\
735 & 100 & 67 & 120 & 795945 & 39807 & 112 & 15222 & 176 & 1.90 & 19113 & 1.93 & 0.165 & 17215 & 0.179\\
745 & 100 & 62 & 113 & 794227 & 29836 & 135 & 15622 & 187 & 1.90 & 19057 & 1.92 & 0.166 & 17300 & 0.181\\
755 & 100 & 40 & 87 & 862698 & 28209 & 136 & 16013 & 195 & 1.89 & 19011 & 1.92 & 0.168 & 17457 & 0.184\\
765 & 100 & 23 & 292 & 902061 & 29092 & 140 & 16279 & 192 & 1.89 & 18954 & 1.91 & 0.170 & 17540 & 0.183\\
775 & 100 & 11 & 490 & 900224 & 17656 & 149 & 16950 & 206 & 1.88 & 18904 & 1.90 & 0.172 & 17623 & 0.182\\
785 & 100 & 8 & 2615 & 824851 & 24910 & 149 & 17385 & 209 & 1.88 & 18874 & 1.90 & 0.169 & 17789 & 0.186\\
795 & 100 & 4 & 2485 & 766795 & 16153 & 145 & 17767 & 204 & 1.88 & 18823 & 1.89 & 0.172 & 17946 & 0.188\\
805 & 100 & 3 & 3138 & 728765 & 14847 & 148 & 17969 & 204 & 1.87 & 18784 & 1.89 & 0.174 & 18057 & 0.188\\
815 & 100 & 1 & 2413 & 650527 & 12647 & 159 & 18643 & 219 & 1.87 & 18744 & 1.88 & 0.176 & 18128 & 0.188\\
825 & 100 & 0 & 2387 & 601167 & 11453 & 168 & 19278 & 227 & 1.86 & 18715 & 1.88 & 0.175 & 18260 & 0.189\\
835 & 100 & 0 & 1852 & 564055 & 9538 & 167 & 19537 & 220 & 1.86 & 18870 & 3.84 & 0.178 & 18414 & 0.191\\
845 & 100 & 0 & 2193 & 531312 & 10982 & 173 & 20025 & 241 & 1.86 & 18643 & 1.87 & 0.178 & 18531 & 0.191\\
855 & 100 & 0 & 2579 & 476522 & 9497 & 178 & 20431 & 244 & 1.85 & 18605 & 1.87 & 0.182 & 18676 & 0.194\\
865 & 100 & 0 & 1949 & 443652 & 7383 & 179 & 20806 & 238 & 1.85 & 18573 & 1.86 & 0.183 & 18799 & 0.193\\
875 & 100 & 0 & 2085 & 413537 & 7209 & 180 & 21352 & 254 & 1.85 & 18531 & 1.86 & 0.182 & 18909 & 0.195\\
\hline
\end{tabular}
\end{table}

\newpage
\begin{table}
\centering
\caption{Full results of the Random SAT algorithm for N=200.}
\begin{tabular}{|r|r|r|r|r|r|r|r|r|r|r|r|r|r|r|}
\hline
600 & 100 & 100 & 51 & 8005 & 336 & 58 & 7891 & 98 & 2.13 & 22009 & 4.12 & 0.131 & 13879 & 0.144\\
610 & 100 & 100 & 39 & 7902 & 175 & 63 & 8275 & 105 & 2.12 & 21694 & 4.11 & 0.133 & 13882 & 0.145\\
620 & 100 & 100 & 51 & 8520 & 228 & 58 & 8273 & 102 & 2.11 & 21187 & 2.13 & 0.133 & 13996 & 0.145\\
630 & 100 & 100 & 51 & 9222 & 316 & 60 & 8619 & 109 & 2.09 & 21671 & 4.09 & 0.134 & 13988 & 0.146\\
640 & 100 & 100 & 44 & 8705 & 198 & 60 & 8945 & 114 & 2.08 & 21164 & 4.08 & 0.134 & 14111 & 0.146\\
650 & 100 & 100 & 45 & 10872 & 483 & 72 & 9292 & 117 & 2.07 & 21062 & 4.06 & 0.135 & 14112 & 0.147\\
660 & 100 & 100 & 45 & 14936 & 1642 & 72 & 9448 & 115 & 2.07 & 20961 & 4.05 & 0.136 & 14213 & 0.150\\
670 & 100 & 100 & 47 & 15419 & 668 & 79 & 9868 & 117 & 2.06 & 20881 & 4.04 & 0.136 & 14267 & 0.147\\
680 & 100 & 100 & 36 & 19914 & 927 & 75 & 10036 & 123 & 2.05 & 20778 & 4.03 & 0.138 & 14331 & 0.152\\
690 & 100 & 100 & 42 & 20593 & 829 & 84 & 10346 & 133 & 2.04 & 20494 & 2.06 & 0.138 & 14418 & 0.150\\
700 & 100 & 100 & 43 & 24919 & 1896 & 84 & 10546 & 129 & 2.03 & 20400 & 2.05 & 0.137 & 14446 & 0.150\\
710 & 100 & 100 & 41 & 29599 & 2107 & 86 & 10939 & 132 & 2.02 & 20321 & 2.05 & 0.140 & 14529 & 0.153\\
720 & 100 & 100 & 47 & 40389 & 3364 & 86 & 11231 & 141 & 2.01 & 20240 & 2.04 & 0.141 & 14682 & 0.153\\
730 & 100 & 100 & 47 & 54922 & 6300 & 88 & 11499 & 136 & 2.01 & 20157 & 2.03 & 0.140 & 14683 & 0.153\\
740 & 100 & 100 & 66 & 74831 & 5181 & 94 & 11908 & 146 & 2.00 & 20088 & 2.02 & 0.140 & 14808 & 0.155\\
750 & 100 & 100 & 42 & 104146 & 7522 & 99 & 12103 & 148 & 1.99 & 20400 & 3.99 & 0.141 & 14918 & 0.155\\
760 & 100 & 100 & 40 & 121366 & 7526 & 99 & 12376 & 144 & 1.98 & 19930 & 2.01 & 0.143 & 14981 & 0.156\\
770 & 100 & 100 & 36 & 189630 & 19232 & 104 & 12951 & 155 & 1.97 & 20061 & 3.96 & 0.143 & 15003 & 0.157\\
780 & 100 & 100 & 87 & 515266 & 41407 & 100 & 13284 & 158 & 1.97 & 20008 & 3.96 & 0.146 & 15111 & 0.158\\
790 & 100 & 100 & 185 & 366716 & 28057 & 107 & 13580 & 164 & 1.96 & 19737 & 1.98 & 0.145 & 15164 & 0.161\\
800 & 100 & 98 & 105 & 704235 & 82068 & 108 & 13855 & 167 & 1.96 & 19678 & 1.98 & 0.147 & 15278 & 0.159\\
810 & 100 & 94 & 196 & 1171447 & 74977 & 113 & 14221 & 168 & 1.95 & 19616 & 1.97 & 0.149 & 15374 & 0.160\\
820 & 100 & 90 & 127 & 1678682 & 93907 & 128 & 14662 & 180 & 1.95 & 19565 & 1.97 & 0.147 & 15459 & 0.160\\
830 & 100 & 85 & 81 & 2167970 & 92260 & 127 & 14899 & 179 & 1.94 & 19710 & 3.93 & 0.150 & 15555 & 0.161\\
840 & 100 & 71 & 153 & 2279300 & 125413 & 128 & 15302 & 186 & 1.93 & 19457 & 1.96 & 0.148 & 15608 & 0.162\\
850 & 100 & 48 & 335 & 2768423 & 81628 & 129 & 15731 & 203 & 1.93 & 19608 & 3.92 & 0.151 & 15738 & 0.162\\
860 & 100 & 38 & 459 & 2818349 & 100712 & 132 & 15943 & 193 & 1.93 & 19346 & 1.94 & 0.152 & 15766 & 0.166\\
870 & 100 & 29 & 704 & 2820201 & 104830 & 139 & 16379 & 195 & 1.92 & 19300 & 1.94 & 0.152 & 15850 & 0.163\\
880 & 100 & 25 & 1208 & 2220243 & 72721 & 135 & 16804 & 192 & 1.92 & 19262 & 1.94 & 0.154 & 16004 & 0.166\\
890 & 100 & 17 & 871 & 2386582 & 63759 & 147 & 17221 & 204 & 1.91 & 19215 & 1.93 & 0.155 & 16076 & 0.167\\
900 & 100 & 3 & 5678 & 2063986 & 47589 & 143 & 17480 & 205 & 1.91 & 19175 & 1.93 & 0.154 & 16141 & 0.167\\
910 & 100 & 4 & 6457 & 1924045 & 56304 & 135 & 17780 & 214 & 1.90 & 19126 & 1.92 & 0.156 & 16268 & 0.168\\
920 & 100 & 1 & 6089 & 1640938 & 65832 & 150 & 18185 & 214 & 1.90 & 19091 & 1.92 & 0.158 & 16367 & 0.168\\
930 & 100 & 1 & 3585 & 1501617 & 41479 & 151 & 18603 & 211 & 1.90 & 19039 & 1.92 & 0.159 & 16471 & 0.169\\
940 & 100 & 0 & 5379 & 1406421 & 37810 & 151 & 18979 & 228 & 1.89 & 19010 & 1.91 & 0.159 & 16529 & 0.171\\
950 & 100 & 0 & 4606 & 1209729 & 26448 & 165 & 19170 & 222 & 1.89 & 18980 & 1.91 & 0.161 & 16667 & 0.172\\
960 & 100 & 0 & 3501 & 1216098 & 35770 & 171 & 19737 & 232 & 1.89 & 18925 & 1.90 & 0.162 & 16693 & 0.173\\
970 & 100 & 0 & 3019 & 1056247 & 36095 & 175 & 20166 & 230 & 1.88 & 18904 & 1.90 & 0.162 & 16807 & 0.174\\
980 & 100 & 0 & 4140 & 920851 & 15625 & 174 & 20613 & 239 & 1.88 & 18885 & 1.90 & 0.164 & 16921 & 0.174\\
990 & 100 & 0 & 3081 & 861910 & 21373 & 180 & 21280 & 242 & 1.88 & 18839 & 1.89 & 0.164 & 17017 & 0.175\\
1000 & 100 & 0 & 3721 & 780571 & 15328 & 182 & 21436 & 247 & 1.87 & 18807 & 1.89 & 0.163 & 17106 & 0.177\\
\hline
\end{tabular}
\end{table}

\newpage
\begin{table}
\centering
\caption{Full results of the Random SAT algorithm for N=225.}
\begin{tabular}{|r|r|r|r|r|r|r|r|r|r|r|r|r|r|r|}
\hline
675 & 25 & 25 & 58 & 2056 & 114 & 65 & 1977 & 93 & 2.18 & 5661 & 4.16 & 0.124 & 3243 & 0.135\\
685 & 25 & 25 & 59 & 2195 & 172 & 69 & 2030 & 103 & 2.17 & 5438 & 2.18 & 0.127 & 3262 & 0.136\\
695 & 25 & 25 & 65 & 2725 & 495 & 66 & 2123 & 106 & 2.16 & 5610 & 4.14 & 0.126 & 3253 & 0.134\\
705 & 25 & 25 & 63 & 3117 & 394 & 60 & 2129 & 104 & 2.15 & 5585 & 4.13 & 0.126 & 3270 & 0.137\\
715 & 25 & 25 & 52 & 2884 & 340 & 79 & 2247 & 112 & 2.13 & 5562 & 4.12 & 0.127 & 3274 & 0.135\\
725 & 25 & 25 & 66 & 3318 & 582 & 70 & 2269 & 108 & 2.13 & 5338 & 2.14 & 0.127 & 3296 & 0.137\\
735 & 25 & 25 & 57 & 5029 & 579 & 77 & 2311 & 108 & 2.12 & 5512 & 4.11 & 0.128 & 3299 & 0.136\\
745 & 25 & 25 & 54 & 4302 & 663 & 77 & 2382 & 115 & 2.10 & 5292 & 2.13 & 0.129 & 3316 & 0.138\\
755 & 25 & 25 & 47 & 4072 & 559 & 74 & 2460 & 115 & 2.10 & 5274 & 2.12 & 0.130 & 3324 & 0.137\\
765 & 25 & 25 & 57 & 4857 & 452 & 78 & 2516 & 125 & 2.09 & 5247 & 2.11 & 0.127 & 3323 & 0.138\\
775 & 25 & 25 & 54 & 5673 & 812 & 80 & 2578 & 121 & 2.08 & 5222 & 2.10 & 0.127 & 3311 & 0.138\\
785 & 25 & 25 & 62 & 5953 & 863 & 92 & 2729 & 130 & 2.07 & 5207 & 2.10 & 0.131 & 3356 & 0.139\\
795 & 25 & 25 & 142 & 11769 & 1099 & 84 & 2706 & 137 & 2.06 & 5185 & 2.08 & 0.130 & 3366 & 0.137\\
805 & 25 & 25 & 52 & 17170 & 2303 & 88 & 2839 & 129 & 2.06 & 5168 & 2.07 & 0.132 & 3379 & 0.138\\
815 & 25 & 25 & 75 & 18034 & 2358 & 87 & 2870 & 132 & 2.05 & 5149 & 2.07 & 0.131 & 3396 & 0.142\\
825 & 25 & 25 & 103 & 32529 & 15195 & 88 & 2896 & 140 & 2.04 & 5129 & 2.06 & 0.130 & 3415 & 0.140\\
835 & 25 & 25 & 61 & 44702 & 14001 & 99 & 2939 & 146 & 2.04 & 5109 & 2.05 & 0.135 & 3424 & 0.141\\
845 & 25 & 25 & 64 & 34479 & 6863 & 108 & 3108 & 146 & 2.03 & 5095 & 2.05 & 0.133 & 3434 & 0.143\\
855 & 25 & 25 & 174 & 51101 & 13828 & 104 & 3108 & 146 & 2.02 & 5076 & 2.04 & 0.134 & 3458 & 0.143\\
865 & 25 & 25 & 127 & 105208 & 30884 & 114 & 3275 & 149 & 2.02 & 5054 & 2.03 & 0.135 & 3466 & 0.141\\
875 & 25 & 25 & 235 & 125632 & 35415 & 108 & 3340 & 150 & 2.01 & 5048 & 2.03 & 0.134 & 3496 & 0.143\\
885 & 25 & 25 & 102 & 137376 & 26304 & 120 & 3437 & 153 & 2.00 & 5027 & 2.02 & 0.135 & 3495 & 0.145\\
895 & 25 & 24 & 249 & 411540 & 177642 & 100 & 3513 & 162 & 2.00 & 5014 & 2.02 & 0.137 & 3513 & 0.145\\
905 & 25 & 25 & 820 & 432532 & 186807 & 119 & 3520 & 158 & 1.99 & 4996 & 2.01 & 0.139 & 3524 & 0.144\\
915 & 25 & 25 & 1949 & 617802 & 137764 & 120 & 3612 & 162 & 1.98 & 4984 & 2.00 & 0.136 & 3539 & 0.146\\
925 & 25 & 21 & 1521 & 1636535 & 209420 & 118 & 3655 & 164 & 1.98 & 4969 & 1.99 & 0.139 & 3567 & 0.146\\
935 & 25 & 20 & 3247 & 1391203 & 170795 & 129 & 3705 & 175 & 1.97 & 4951 & 1.99 & 0.139 & 3579 & 0.147\\
945 & 25 & 19 & 3128 & 1291132 & 166016 & 129 & 3759 & 178 & 1.97 & 4945 & 1.98 & 0.140 & 3606 & 0.148\\
955 & 25 & 13 & 3834 & 2271068 & 245304 & 138 & 3917 & 176 & 1.97 & 4933 & 1.98 & 0.140 & 3614 & 0.151\\
965 & 25 & 11 & 1616 & 1918131 & 148155 & 126 & 4010 & 178 & 1.96 & 4919 & 1.98 & 0.142 & 3645 & 0.151\\
975 & 25 & 7 & 603 & 1856267 & 138588 & 137 & 4112 & 199 & 1.96 & 4906 & 1.97 & 0.143 & 3638 & 0.148\\
985 & 25 & 3 & 10011 & 2096434 & 192684 & 144 & 4143 & 189 & 1.95 & 4892 & 1.97 & 0.142 & 3653 & 0.151\\
995 & 25 & 7 & 8391 & 2047771 & 142322 & 150 & 4212 & 202 & 1.95 & 4881 & 1.96 & 0.143 & 3661 & 0.152\\
1005 & 25 & 2 & 5771 & 1511449 & 157374 & 151 & 4313 & 197 & 1.94 & 4873 & 1.96 & 0.145 & 3705 & 0.153\\
1015 & 25 & 1 & 20027 & 1506162 & 108728 & 156 & 4403 & 192 & 1.94 & 4860 & 1.95 & 0.145 & 3719 & 0.154\\
1025 & 25 & 1 & 28137 & 1398087 & 106039 & 153 & 4476 & 197 & 1.93 & 4852 & 1.95 & 0.144 & 3728 & 0.154\\
1035 & 25 & 1 & 20890 & 1268619 & 113910 & 159 & 4502 & 204 & 1.93 & 4842 & 1.94 & 0.143 & 3743 & 0.153\\
1045 & 25 & 0 & 15987 & 1074700 & 87185 & 158 & 4568 & 211 & 1.92 & 4827 & 1.94 & 0.147 & 3764 & 0.155\\
1055 & 25 & 0 & 9698 & 813706 & 63433 & 167 & 4806 & 229 & 1.92 & 4826 & 1.94 & 0.147 & 3778 & 0.155\\
1065 & 25 & 0 & 12595 & 831258 & 75407 & 166 & 4903 & 221 & 1.92 & 4814 & 1.93 & 0.148 & 3792 & 0.156\\
1075 & 25 & 0 & 14172 & 715952 & 80468 & 158 & 4879 & 217 & 1.91 & 4805 & 1.93 & 0.148 & 3824 & 0.158\\
1085 & 25 & 0 & 10164 & 622022 & 85008 & 172 & 5055 & 222 & 1.91 & 4800 & 1.93 & 0.151 & 3834 & 0.156\\
1095 & 25 & 0 & 9596 & 575054 & 36154 & 186 & 5152 & 234 & 1.91 & 4792 & 1.92 & 0.149 & 3862 & 0.158\\
1105 & 25 & 0 & 5925 & 515638 & 40923 & 178 & 5298 & 239 & 1.91 & 4782 & 1.92 & 0.149 & 3867 & 0.159\\
1115 & 25 & 0 & 8207 & 460458 & 34261 & 188 & 5272 & 241 & 1.90 & 4773 & 1.92 & 0.150 & 3890 & 0.163\\
1125 & 25 & 0 & 7249 & 486975 & 44923 & 189 & 5448 & 249 & 1.90 & 4765 & 1.91 & 0.153 & 3919 & 0.161\\
\hline
\end{tabular}
\end{table}

\newpage
\begin{table}
\centering
\caption{Full results of the Balanced SAT algorithm for N=175.}
\begin{tabular}{|r|r|r|r|r|r|r|r|r|r|r|r|r|r|r|}
\hline
525 & 100 & 100 & 38 & 62211 & 2594 & 0 & 62 & 4 & 2.02 & 20292 & 2.04 & 0.129 & 13613 & 0.141\\
535 & 100 & 100 & 38 & 100213 & 4820 & 0 & 67 & 4 & 2.01 & 20174 & 2.02 & 0.132 & 13728 & 0.143\\
545 & 100 & 100 & 79 & 164864 & 14989 & 0 & 66 & 4 & 2.00 & 20046 & 2.01 & 0.133 & 13803 & 0.144\\
555 & 100 & 100 & 80 & 394793 & 20457 & 0 & 76 & 3 & 1.99 & 19933 & 2.00 & 0.131 & 13915 & 0.145\\
565 & 100 & 100 & 72 & 976117 & 94456 & 0 & 51 & 2 & 1.97 & 19809 & 1.99 & 0.133 & 13957 & 0.148\\
575 & 100 & 100 & 159 & 2709697 & 229764 & 0 & 60 & 3 & 1.97 & 19705 & 1.98 & 0.136 & 14096 & 0.148\\
585 & 100 & 100 & 735 & 10110047 & 596479 & 0 & 69 & 3 & 1.95 & 19604 & 1.97 & 0.133 & 14171 & 0.148\\
595 & 100 & 97 & 640 & 27515142 & 2944991 & 0 & 73 & 3 & 1.95 & 19511 & 1.96 & 0.137 & 14302 & 0.150\\
605 & 100 & 83 & 505 & 49015355 & 2761113 & 0 & 73 & 4 & 1.94 & 19431 & 1.95 & 0.141 & 14451 & 0.151\\
615 & 100 & 51 & 1524 & 73100116 & 1636515 & 0 & 82 & 3 & 1.93 & 19362 & 1.94 & 0.140 & 14590 & 0.152\\
625 & 100 & 28 & 2123 & 77268614 & 1265079 & 0 & 79 & 3 & 1.92 & 19291 & 1.93 & 0.141 & 14671 & 0.154\\
635 & 100 & 11 & 4368 & 78283199 & 1095609 & 0 & 78 & 3 & 1.92 & 19201 & 1.93 & 0.141 & 14800 & 0.154\\
645 & 100 & 2 & 407930 & 68378942 & 991447 & 0 & 78 & 3 & 1.91 & 19116 & 1.92 & 0.144 & 14860 & 0.154\\
655 & 100 & 3 & 109013 & 49640591 & 675064 & 0 & 77 & 4 & 1.90 & 19082 & 1.91 & 0.145 & 15023 & 0.155\\
665 & 100 & 0 & 322458 & 42706631 & 579850 & 0 & 91 & 3 & 1.90 & 19000 & 1.90 & 0.148 & 15139 & 0.156\\
675 & 100 & 0 & 283517 & 37265554 & 484737 & 0 & 77 & 3 & 1.89 & 18942 & 1.90 & 0.147 & 15294 & 0.160\\
685 & 100 & 0 & 263031 & 33057198 & 425587 & 0 & 94 & 4 & 1.89 & 18900 & 1.89 & 0.148 & 15422 & 0.160\\
695 & 100 & 0 & 234532 & 30268101 & 373855 & 0 & 102 & 4 & 1.88 & 18824 & 1.89 & 0.151 & 15518 & 0.159\\
705 & 100 & 0 & 197172 & 26637122 & 330557 & 0 & 125 & 4 & 1.88 & 18800 & 1.88 & 0.151 & 15701 & 0.163\\
715 & 100 & 0 & 155588 & 23258828 & 294058 & 0 & 131 & 4 & 1.87 & 18743 & 1.88 & 0.154 & 15802 & 0.164\\
725 & 100 & 0 & 118915 & 20551706 & 267380 & 0 & 109 & 4 & 1.87 & 18700 & 1.87 & 0.154 & 15996 & 0.167\\
735 & 100 & 0 & 91289 & 18750455 & 243652 & 0 & 138 & 5 & 1.86 & 18696 & 1.87 & 0.156 & 16119 & 0.169\\
745 & 100 & 0 & 88356 & 15771666 & 219329 & 0 & 109 & 4 & 1.86 & 18600 & 1.86 & 0.159 & 16293 & 0.167\\
755 & 100 & 0 & 70045 & 14493008 & 213955 & 0 & 125 & 4 & 1.86 & 18600 & 1.86 & 0.159 & 16381 & 0.169\\
765 & 100 & 0 & 68095 & 11546446 & 187105 & 0 & 121 & 3 & 1.85 & 18555 & 1.86 & 0.160 & 16542 & 0.172\\
775 & 100 & 0 & 50243 & 8782812 & 139058 & 0 & 125 & 4 & 1.85 & 18500 & 1.85 & 0.162 & 16654 & 0.172\\
785 & 100 & 0 & 39062 & 7598549 & 141292 & 0 & 122 & 5 & 1.85 & 18500 & 1.85 & 0.162 & 16804 & 0.172\\
795 & 100 & 0 & 37432 & 6058963 & 106575 & 0 & 138 & 5 & 1.84 & 18491 & 1.85 & 0.167 & 17001 & 0.174\\
805 & 100 & 0 & 28661 & 5200819 & 103522 & 0 & 136 & 4 & 1.84 & 18400 & 1.84 & 0.166 & 17126 & 0.176\\
815 & 100 & 0 & 26623 & 4464272 & 83275 & 0 & 120 & 5 & 1.84 & 18400 & 1.84 & 0.168 & 17287 & 0.179\\
825 & 100 & 0 & 19140 & 3588886 & 60989 & 0 & 167 & 6 & 1.84 & 18400 & 1.84 & 0.171 & 17439 & 0.178\\
835 & 100 & 0 & 15780 & 2956467 & 60339 & 0 & 130 & 5 & 1.83 & 18331 & 1.84 & 0.172 & 17617 & 0.180\\
845 & 100 & 0 & 14869 & 2566463 & 43399 & 0 & 142 & 5 & 1.83 & 18300 & 1.83 & 0.171 & 17755 & 0.181\\
855 & 100 & 0 & 13580 & 2271585 & 41845 & 0 & 169 & 5 & 1.83 & 18300 & 1.83 & 0.176 & 17909 & 0.183\\
865 & 100 & 0 & 12249 & 2011425 & 35015 & 0 & 144 & 5 & 1.83 & 18300 & 1.83 & 0.174 & 18091 & 0.186\\
875 & 100 & 0 & 11322 & 1834440 & 29685 & 0 & 202 & 7 & 1.82 & 18232 & 1.83 & 0.178 & 18234 & 0.187\\
\hline
\end{tabular}
\end{table}

\newpage
\begin{table}
\centering
\caption{Full results of the Balanced SAT algorithm for N=200.}
\begin{tabular}{|r|r|r|r|r|r|r|r|r|r|r|r|r|r|r|}
\hline
600 & 100 & 100 & 60 & 74744 & 5129 & 0 & 51 & 3 & 2.08 & 20866 & 2.09 & 0.121 & 12617 & 0.132\\
610 & 100 & 100 & 63 & 136542 & 11104 & 0 & 46 & 2 & 2.07 & 20748 & 2.08 & 0.121 & 12712 & 0.133\\
620 & 100 & 100 & 76 & 267365 & 17108 & 0 & 64 & 3 & 2.06 & 20622 & 2.07 & 0.122 & 12724 & 0.132\\
630 & 100 & 100 & 132 & 462105 & 15857 & 0 & 57 & 4 & 2.04 & 20504 & 2.06 & 0.123 & 12841 & 0.135\\
640 & 100 & 100 & 213 & 913926 & 40134 & 0 & 56 & 3 & 2.03 & 20401 & 2.05 & 0.123 & 12884 & 0.137\\
650 & 100 & 100 & 386 & 2730960 & 209636 & 0 & 48 & 2 & 2.02 & 20301 & 2.04 & 0.123 & 12976 & 0.135\\
660 & 100 & 100 & 2126 & 10491079 & 445113 & 0 & 60 & 3 & 2.01 & 20192 & 2.02 & 0.125 & 12944 & 0.135\\
670 & 100 & 100 & 959 & 25412661 & 2942769 & 0 & 58 & 4 & 2.00 & 20093 & 2.02 & 0.125 & 13048 & 0.134\\
680 & 100 & 99 & 3144 & 87341132 & 14431020 & 0 & 76 & 4 & 1.99 & 20001 & 2.01 & 0.126 & 13191 & 0.140\\
690 & 100 & 88 & 4753 & 268528080 & 17891392 & 0 & 76 & 3 & 1.98 & 19903 & 2.00 & 0.126 & 13209 & 0.137\\
700 & 100 & 70 & 958 & 413088488 & 13564611 & 0 & 59 & 4 & 1.98 & 19817 & 1.99 & 0.129 & 13283 & 0.138\\
710 & 100 & 37 & 218652 & 625956584 & 13015449 & 0 & 63 & 3 & 1.97 & 19740 & 1.98 & 0.126 & 13358 & 0.139\\
720 & 100 & 18 & 4620 & 605551901 & 10477146 & 0 & 64 & 3 & 1.96 & 19678 & 1.97 & 0.130 & 13431 & 0.141\\
730 & 100 & 6 & 379963 & 583539700 & 8992637 & 0 & 60 & 3 & 1.95 & 19596 & 1.96 & 0.130 & 13495 & 0.140\\
740 & 100 & 4 & 203102 & 364610273 & 6139167 & 0 & 82 & 3 & 1.95 & 19504 & 1.96 & 0.131 & 13617 & 0.142\\
750 & 100 & 1 & 255836 & 240138858 & 4156333 & 0 & 81 & 3 & 1.94 & 19450 & 1.95 & 0.131 & 13704 & 0.143\\
760 & 100 & 0 & 868907 & 163639814 & 2735728 & 0 & 70 & 3 & 1.93 & 19399 & 1.94 & 0.134 & 13777 & 0.142\\
770 & 100 & 0 & 772500 & 120798167 & 2326081 & 0 & 67 & 3 & 1.93 & 19312 & 1.94 & 0.134 & 13902 & 0.144\\
780 & 100 & 0 & 641112 & 100678962 & 1366955 & 0 & 78 & 3 & 1.92 & 19286 & 1.93 & 0.133 & 13970 & 0.144\\
790 & 100 & 0 & 621077 & 87999516 & 1191000 & 0 & 67 & 3 & 1.92 & 19201 & 1.93 & 0.134 & 14054 & 0.147\\
800 & 100 & 0 & 587903 & 78527861 & 1084444 & 0 & 115 & 3 & 1.91 & 19163 & 1.92 & 0.136 & 14127 & 0.146\\
810 & 100 & 0 & 483230 & 65370428 & 897547 & 0 & 102 & 3 & 1.91 & 19101 & 1.92 & 0.138 & 14261 & 0.148\\
820 & 100 & 0 & 390575 & 54384554 & 738202 & 0 & 103 & 4 & 1.90 & 19088 & 1.91 & 0.139 & 14369 & 0.149\\
830 & 100 & 0 & 319832 & 47786620 & 742219 & 0 & 100 & 5 & 1.90 & 19001 & 1.91 & 0.140 & 14457 & 0.149\\
840 & 100 & 0 & 317547 & 44402306 & 674473 & 0 & 82 & 4 & 1.89 & 18998 & 1.90 & 0.142 & 14556 & 0.151\\
850 & 100 & 0 & 321862 & 40718055 & 525031 & 0 & 107 & 4 & 1.89 & 18916 & 1.90 & 0.139 & 14671 & 0.151\\
860 & 100 & 0 & 299190 & 37648681 & 478997 & 0 & 105 & 4 & 1.89 & 18900 & 1.89 & 0.144 & 14781 & 0.153\\
870 & 100 & 0 & 261517 & 35122229 & 432253 & 0 & 110 & 4 & 1.88 & 18871 & 1.89 & 0.144 & 14865 & 0.154\\
880 & 100 & 0 & 187001 & 30046977 & 375641 & 0 & 118 & 4 & 1.88 & 18800 & 1.88 & 0.146 & 14995 & 0.155\\
890 & 100 & 0 & 202904 & 27422982 & 327054 & 0 & 130 & 5 & 1.88 & 18800 & 1.88 & 0.147 & 15068 & 0.155\\
900 & 100 & 0 & 163013 & 25148598 & 317540 & 0 & 112 & 4 & 1.87 & 18781 & 1.88 & 0.148 & 15204 & 0.160\\
910 & 100 & 0 & 130842 & 23338186 & 300293 & 0 & 103 & 4 & 1.87 & 18700 & 1.87 & 0.149 & 15305 & 0.158\\
920 & 100 & 0 & 133906 & 21502083 & 266971 & 0 & 110 & 6 & 1.87 & 18700 & 1.87 & 0.152 & 15447 & 0.159\\
930 & 100 & 0 & 125661 & 20442322 & 254168 & 0 & 121 & 4 & 1.87 & 18700 & 1.87 & 0.151 & 15507 & 0.159\\
940 & 100 & 0 & 109753 & 17865252 & 243029 & 0 & 126 & 4 & 1.86 & 18610 & 1.87 & 0.152 & 15625 & 0.161\\
950 & 100 & 0 & 87633 & 15355302 & 229197 & 0 & 137 & 4 & 1.86 & 18600 & 1.86 & 0.152 & 15749 & 0.162\\
960 & 100 & 0 & 43686 & 12666829 & 194142 & 0 & 138 & 5 & 1.86 & 18600 & 1.86 & 0.155 & 15875 & 0.164\\
970 & 100 & 0 & 51827 & 11240234 & 186997 & 0 & 116 & 5 & 1.85 & 18598 & 1.86 & 0.157 & 15988 & 0.164\\
980 & 100 & 0 & 46386 & 10334539 & 199723 & 0 & 131 & 5 & 1.85 & 18519 & 1.86 & 0.157 & 16096 & 0.165\\
990 & 100 & 0 & 44074 & 9251583 & 154012 & 0 & 125 & 4 & 1.85 & 18500 & 1.85 & 0.158 & 16198 & 0.166\\
1000 & 100 & 0 & 37626 & 8387776 & 154786 & 0 & 172 & 5 & 1.85 & 18500 & 1.85 & 0.159 & 16317 & 0.169\\
\hline
\end{tabular}
\end{table}

\newpage
\begin{table}
\centering
\caption{Full results of the Balanced SAT algorithm for N=225.}
\begin{tabular}{|r|r|r|r|r|r|r|r|r|r|r|r|r|r|r|}
\hline
675 & 25 & 25 & 209 & 28358 & 3834 & 0 & 22 & 3 & 2.13 & 5345 & 2.14 & 0.114 & 2952 & 0.121\\
685 & 25 & 25 & 146 & 48531 & 6998 & 0 & 35 & 4 & 2.12 & 5324 & 2.13 & 0.115 & 2972 & 0.123\\
695 & 25 & 25 & 304 & 62256 & 15549 & 0 & 20 & 2 & 2.11 & 5290 & 2.12 & 0.114 & 2999 & 0.124\\
705 & 25 & 25 & 397 & 156528 & 21884 & 0 & 15 & 2 & 2.10 & 5265 & 2.11 & 0.116 & 2987 & 0.123\\
715 & 25 & 25 & 415 & 317493 & 48694 & 0 & 19 & 3 & 2.09 & 5240 & 2.10 & 0.117 & 3017 & 0.126\\
725 & 25 & 25 & 2003 & 499112 & 49577 & 0 & 29 & 3 & 2.08 & 5208 & 2.09 & 0.117 & 2986 & 0.123\\
735 & 25 & 25 & 1903 & 3374264 & 1167695 & 0 & 24 & 4 & 2.07 & 5182 & 2.08 & 0.116 & 3005 & 0.124\\
745 & 25 & 25 & 21436 & 9952167 & 2311298 & 0 & 24 & 4 & 2.06 & 5153 & 2.07 & 0.116 & 3019 & 0.127\\
755 & 25 & 25 & 31045 & 33682937 & 7455734 & 0 & 32 & 4 & 2.05 & 5135 & 2.06 & 0.119 & 3040 & 0.124\\
765 & 25 & 25 & 47021 & 50270191 & 20427104 & 0 & 30 & 3 & 2.04 & 5112 & 2.05 & 0.118 & 3045 & 0.127\\
775 & 25 & 24 & 147821 & 262667419 & 65268387 & 0 & 33 & 4 & 2.03 & 5096 & 2.04 & 0.119 & 3070 & 0.128\\
785 & 25 & 19 & 53865 & 437422958 & 49715216 & 0 & 23 & 3 & 2.02 & 5071 & 2.03 & 0.119 & 3082 & 0.129\\
795 & 25 & 12 & 558206 & 515490231 & 39856053 & 0 & 33 & 3 & 2.02 & 5052 & 2.03 & 0.121 & 3109 & 0.127\\
805 & 25 & 9 & 272878 & 617793436 & 44995962 & 0 & 21 & 3 & 2.01 & 5026 & 2.02 & 0.120 & 3097 & 0.127\\
815 & 25 & 4 & 1219470 & 587978520 & 34168620 & 0 & 28 & 3 & 2.00 & 5005 & 2.01 & 0.120 & 3110 & 0.127\\
825 & 25 & 2 & 4794497 & 566239510 & 28375650 & 0 & 38 & 5 & 1.99 & 4994 & 2.00 & 0.121 & 3131 & 0.128\\
835 & 25 & 0 & 10057838 & 421277510 & 22489592 & 0 & 31 & 3 & 1.99 & 4975 & 1.99 & 0.122 & 3142 & 0.130\\
845 & 25 & 0 & 9622275 & 336232127 & 17730908 & 0 & 22 & 2 & 1.98 & 4952 & 1.99 & 0.124 & 3171 & 0.130\\
855 & 25 & 0 & 7667378 & 272971958 & 16380329 & 0 & 40 & 6 & 1.97 & 4948 & 1.98 & 0.124 & 3186 & 0.131\\
865 & 25 & 0 & 5018906 & 210076932 & 11378817 & 0 & 37 & 5 & 1.97 & 4927 & 1.98 & 0.124 & 3205 & 0.134\\
875 & 25 & 0 & 4008141 & 175674993 & 10820652 & 0 & 37 & 4 & 1.96 & 4912 & 1.97 & 0.126 & 3222 & 0.132\\
885 & 25 & 0 & 2974408 & 147699827 & 9228776 & 0 & 40 & 4 & 1.96 & 4900 & 1.96 & 0.126 & 3247 & 0.133\\
895 & 25 & 0 & 3025087 & 128306110 & 8337797 & 0 & 34 & 3 & 1.95 & 4879 & 1.96 & 0.126 & 3259 & 0.134\\
905 & 25 & 0 & 1895433 & 94539106 & 5306052 & 0 & 43 & 5 & 1.94 & 4872 & 1.95 & 0.127 & 3258 & 0.133\\
915 & 25 & 0 & 1681623 & 72389634 & 4248709 & 0 & 47 & 5 & 1.94 & 4856 & 1.95 & 0.129 & 3294 & 0.137\\
925 & 25 & 0 & 1356969 & 51835486 & 3698778 & 0 & 38 & 4 & 1.93 & 4849 & 1.94 & 0.129 & 3309 & 0.136\\
935 & 25 & 0 & 1062796 & 40665858 & 3185711 & 0 & 39 & 4 & 1.93 & 4826 & 1.94 & 0.130 & 3325 & 0.137\\
945 & 25 & 0 & 983563 & 32042746 & 1704670 & 1 & 58 & 5 & 1.93 & 4825 & 1.93 & 0.129 & 3331 & 0.138\\
955 & 25 & 0 & 816725 & 28099763 & 1689906 & 0 & 46 & 6 & 1.92 & 4807 & 1.93 & 0.131 & 3355 & 0.137\\
965 & 25 & 0 & 791516 & 26757741 & 1500555 & 0 & 43 & 5 & 1.92 & 4800 & 1.92 & 0.133 & 3387 & 0.139\\
975 & 25 & 0 & 711557 & 24236465 & 1266714 & 0 & 45 & 5 & 1.91 & 4794 & 1.92 & 0.132 & 3395 & 0.140\\
985 & 25 & 0 & 563889 & 19713949 & 1002265 & 0 & 52 & 5 & 1.91 & 4775 & 1.91 & 0.134 & 3420 & 0.140\\
995 & 25 & 0 & 479524 & 16308456 & 830998 & 0 & 62 & 5 & 1.91 & 4775 & 1.91 & 0.134 & 3433 & 0.141\\
1005 & 25 & 0 & 454633 & 14960311 & 815451 & 0 & 60 & 4 & 1.90 & 4762 & 1.91 & 0.135 & 3454 & 0.141\\
1015 & 25 & 0 & 412768 & 13475652 & 673255 & 0 & 52 & 5 & 1.90 & 4750 & 1.90 & 0.133 & 3478 & 0.143\\
1025 & 25 & 0 & 394527 & 12194125 & 554821 & 0 & 42 & 4 & 1.90 & 4750 & 1.90 & 0.137 & 3503 & 0.143\\
1035 & 25 & 0 & 339646 & 10969882 & 510657 & 0 & 39 & 4 & 1.89 & 4742 & 1.90 & 0.138 & 3516 & 0.144\\
1045 & 25 & 0 & 338860 & 10427428 & 498608 & 0 & 63 & 7 & 1.89 & 4725 & 1.89 & 0.138 & 3534 & 0.145\\
1055 & 25 & 0 & 283251 & 9401119 & 493375 & 0 & 62 & 6 & 1.89 & 4725 & 1.89 & 0.139 & 3577 & 0.146\\
1065 & 25 & 0 & 271950 & 8891702 & 446883 & 0 & 62 & 6 & 1.88 & 4724 & 1.89 & 0.141 & 3606 & 0.148\\
1075 & 25 & 0 & 285650 & 8330238 & 401701 & 0 & 69 & 6 & 1.88 & 4705 & 1.89 & 0.141 & 3600 & 0.148\\
1085 & 25 & 0 & 258551 & 7828755 & 379711 & 0 & 57 & 6 & 1.88 & 4700 & 1.88 & 0.141 & 3630 & 0.148\\
1095 & 25 & 0 & 200311 & 7219457 & 339582 & 0 & 55 & 6 & 1.88 & 4700 & 1.88 & 0.143 & 3648 & 0.148\\
1105 & 25 & 0 & 219805 & 6912573 & 337282 & 0 & 53 & 6 & 1.87 & 4699 & 1.88 & 0.144 & 3660 & 0.150\\
1115 & 25 & 0 & 176719 & 6500832 & 306719 & 1 & 69 & 6 & 1.87 & 4678 & 1.88 & 0.146 & 3694 & 0.151\\
1125 & 25 & 0 & 202431 & 6321020 & 318091 & 0 & 74 & 7 & 1.87 & 4675 & 1.87 & 0.146 & 3716 & 0.152\\
\hline
\end{tabular}
\end{table}

\newpage
\begin{table}
\centering
\caption{Full results of the No-Triangle SAT algorithm for N=175.}
\begin{tabular}{|r|r|r|r|r|r|r|r|r|r|r|r|r|r|r|}
\hline
525 & 100 & 100 & 37 & 15250 & 712 & 0 & 76 & 5 & 1.98 & 19913 & 2.00 & 0.061 & 6214 & 0.063\\
535 & 100 & 100 & 35 & 17553 & 653 & 0 & 84 & 3 & 1.98 & 19804 & 1.99 & 0.060 & 6099 & 0.062\\
545 & 100 & 100 & 47 & 23402 & 1371 & 0 & 96 & 3 & 1.97 & 19701 & 1.98 & 0.059 & 6006 & 0.061\\
555 & 100 & 100 & 43 & 32386 & 1534 & 0 & 79 & 3 & 1.96 & 19601 & 1.97 & 0.058 & 5947 & 0.061\\
565 & 100 & 100 & 42 & 40186 & 3952 & 0 & 85 & 4 & 1.95 & 19501 & 1.96 & 0.058 & 5882 & 0.060\\
575 & 100 & 100 & 39 & 58717 & 1555 & 0 & 88 & 3 & 1.94 & 19400 & 1.94 & 0.057 & 5849 & 0.060\\
585 & 100 & 100 & 47 & 100915 & 5985 & 0 & 79 & 4 & 1.93 & 19310 & 1.94 & 0.057 & 5916 & 0.060\\
595 & 100 & 100 & 43 & 106317 & 5623 & 0 & 104 & 4 & 1.92 & 19238 & 1.93 & 0.057 & 5860 & 0.061\\
605 & 100 & 100 & 102 & 164526 & 5687 & 0 & 90 & 4 & 1.91 & 19192 & 1.92 & 0.057 & 5815 & 0.060\\
615 & 100 & 100 & 64 & 276891 & 13203 & 0 & 83 & 3 & 1.91 & 19100 & 1.91 & 0.056 & 5786 & 0.059\\
625 & 100 & 100 & 45 & 388124 & 23571 & 0 & 94 & 3 & 1.90 & 19023 & 1.91 & 0.056 & 5766 & 0.059\\
635 & 100 & 100 & 47 & 795201 & 39282 & 0 & 102 & 4 & 1.89 & 18999 & 1.90 & 0.056 & 5800 & 0.059\\
645 & 100 & 100 & 105 & 1072264 & 64565 & 0 & 117 & 5 & 1.89 & 18908 & 1.90 & 0.058 & 5891 & 0.060\\
655 & 100 & 100 & 225 & 1776136 & 107172 & 0 & 125 & 4 & 1.88 & 18895 & 1.89 & 0.057 & 5873 & 0.061\\
665 & 100 & 100 & 320 & 3252788 & 232143 & 0 & 129 & 4 & 1.88 & 18800 & 1.88 & 0.057 & 5869 & 0.060\\
675 & 100 & 100 & 409 & 6499525 & 404988 & 0 & 107 & 5 & 1.87 & 18799 & 1.88 & 0.058 & 5877 & 0.060\\
685 & 100 & 100 & 1134 & 13975071 & 2984858 & 0 & 121 & 4 & 1.87 & 18703 & 1.88 & 0.058 & 5906 & 0.060\\
695 & 100 & 92 & 851 & 29184354 & 2433899 & 0 & 118 & 4 & 1.87 & 18700 & 1.87 & 0.058 & 5982 & 0.062\\
705 & 100 & 73 & 958 & 51389353 & 1913059 & 0 & 150 & 6 & 1.86 & 18686 & 1.87 & 0.059 & 6103 & 0.062\\
715 & 100 & 53 & 665 & 61205027 & 1165890 & 0 & 174 & 5 & 1.86 & 18600 & 1.86 & 0.060 & 6129 & 0.063\\
725 & 100 & 36 & 4078 & 61913402 & 1064983 & 0 & 177 & 5 & 1.86 & 18600 & 1.86 & 0.060 & 6141 & 0.063\\
735 & 100 & 14 & 5328 & 66645732 & 1010250 & 0 & 169 & 5 & 1.85 & 18573 & 1.86 & 0.060 & 6180 & 0.064\\
745 & 100 & 12 & 148635 & 58297842 & 799695 & 0 & 163 & 5 & 1.85 & 18500 & 1.85 & 0.061 & 6253 & 0.064\\
755 & 100 & 3 & 154118 & 53801432 & 689309 & 0 & 158 & 5 & 1.85 & 18500 & 1.85 & 0.062 & 6394 & 0.066\\
765 & 100 & 1 & 128819 & 46673669 & 578716 & 0 & 157 & 5 & 1.85 & 18500 & 1.85 & 0.063 & 6495 & 0.066\\
775 & 100 & 1 & 341182 & 42206118 & 508810 & 0 & 154 & 5 & 1.84 & 18408 & 1.85 & 0.064 & 6537 & 0.068\\
785 & 100 & 0 & 317253 & 38202177 & 469590 & 0 & 189 & 6 & 1.84 & 18400 & 1.84 & 0.064 & 6586 & 0.068\\
795 & 100 & 1 & 303081 & 34972710 & 407504 & 0 & 171 & 6 & 1.84 & 18400 & 1.84 & 0.065 & 6671 & 0.068\\
805 & 100 & 0 & 260478 & 32099887 & 385377 & 0 & 163 & 5 & 1.84 & 18400 & 1.84 & 0.066 & 6795 & 0.070\\
815 & 100 & 0 & 240068 & 29659596 & 345337 & 0 & 162 & 6 & 1.83 & 18389 & 1.84 & 0.068 & 6964 & 0.071\\
825 & 100 & 0 & 187611 & 26867156 & 306272 & 0 & 170 & 8 & 1.83 & 18300 & 1.83 & 0.069 & 7073 & 0.073\\
835 & 100 & 0 & 170978 & 24218808 & 293286 & 0 & 187 & 5 & 1.83 & 18300 & 1.83 & 0.070 & 7147 & 0.073\\
845 & 100 & 0 & 170620 & 22976290 & 265154 & 0 & 175 & 4 & 1.83 & 18300 & 1.83 & 0.070 & 7234 & 0.074\\
855 & 100 & 0 & 117120 & 20380808 & 255330 & 0 & 216 & 6 & 1.83 & 18300 & 1.83 & 0.071 & 7344 & 0.076\\
865 & 100 & 0 & 119202 & 18756778 & 236289 & 0 & 181 & 5 & 1.82 & 18203 & 1.83 & 0.073 & 7475 & 0.076\\
875 & 100 & 0 & 78535 & 16117067 & 222456 & 0 & 246 & 8 & 1.82 & 18200 & 1.82 & 0.076 & 7726 & 0.079\\
\hline
\end{tabular}
\end{table}

\newpage
\begin{table}
\centering
\caption{Full results of the No-Triangle SAT algorithm for N=200.}
\begin{tabular}{|r|r|r|r|r|r|r|r|r|r|r|r|r|r|r|}
\hline
600 & 100 & 100 & 52 & 23775 & 830 & 0 & 63 & 3 & 2.04 & 20486 & 2.05 & 0.060 & 6054 & 0.062\\
610 & 100 & 100 & 42 & 30961 & 1253 & 0 & 76 & 4 & 2.03 & 20373 & 2.04 & 0.059 & 5932 & 0.060\\
620 & 100 & 100 & 41 & 36740 & 1813 & 0 & 73 & 3 & 2.02 & 20247 & 2.03 & 0.057 & 5837 & 0.059\\
630 & 100 & 100 & 48 & 54935 & 3465 & 0 & 74 & 3 & 2.01 & 20141 & 2.02 & 0.057 & 5740 & 0.059\\
640 & 100 & 100 & 51 & 67918 & 3654 & 0 & 73 & 4 & 2.00 & 20036 & 2.01 & 0.056 & 5664 & 0.058\\
650 & 100 & 100 & 41 & 102134 & 5155 & 0 & 90 & 3 & 1.99 & 19932 & 2.00 & 0.055 & 5590 & 0.057\\
660 & 100 & 100 & 105 & 183465 & 9505 & 0 & 85 & 4 & 1.98 & 19836 & 1.99 & 0.055 & 5557 & 0.056\\
670 & 100 & 100 & 36 & 234516 & 14067 & 0 & 86 & 3 & 1.97 & 19775 & 1.98 & 0.055 & 5570 & 0.057\\
680 & 100 & 100 & 77 & 381684 & 16615 & 0 & 89 & 5 & 1.96 & 19693 & 1.97 & 0.054 & 5507 & 0.056\\
690 & 100 & 100 & 65 & 497286 & 29871 & 0 & 93 & 4 & 1.96 & 19600 & 1.96 & 0.053 & 5446 & 0.056\\
700 & 100 & 100 & 197 & 664614 & 30768 & 0 & 96 & 4 & 1.95 & 19507 & 1.96 & 0.053 & 5394 & 0.055\\
710 & 100 & 100 & 312 & 1111412 & 46512 & 0 & 93 & 4 & 1.94 & 19461 & 1.95 & 0.053 & 5359 & 0.055\\
720 & 100 & 100 & 328 & 2073675 & 132703 & 0 & 103 & 4 & 1.93 & 19399 & 1.94 & 0.052 & 5328 & 0.054\\
730 & 100 & 100 & 55 & 4562244 & 348829 & 0 & 89 & 3 & 1.93 & 19303 & 1.94 & 0.053 & 5352 & 0.055\\
740 & 100 & 100 & 112 & 6151038 & 475253 & 0 & 112 & 4 & 1.92 & 19287 & 1.93 & 0.053 & 5389 & 0.055\\
750 & 100 & 100 & 455 & 11344065 & 472765 & 0 & 112 & 4 & 1.92 & 19200 & 1.92 & 0.052 & 5345 & 0.055\\
760 & 100 & 100 & 1660 & 20156398 & 1043815 & 0 & 103 & 4 & 1.91 & 19175 & 1.92 & 0.051 & 5319 & 0.055\\
770 & 100 & 100 & 2484 & 45560823 & 4152091 & 0 & 105 & 6 & 1.91 & 19100 & 1.91 & 0.052 & 5298 & 0.054\\
780 & 100 & 98 & 5214 & 125347593 & 21975427 & 0 & 98 & 3 & 1.90 & 19073 & 1.91 & 0.052 & 5305 & 0.055\\
790 & 100 & 79 & 2517 & 463156885 & 22012457 & 0 & 104 & 4 & 1.90 & 19000 & 1.90 & 0.052 & 5314 & 0.055\\
800 & 100 & 46 & 1700 & 888062820 & 20976664 & 0 & 140 & 5 & 1.89 & 18999 & 1.90 & 0.053 & 5431 & 0.055\\
810 & 100 & 31 & 35029 & 748102234 & 15553241 & 0 & 156 & 6 & 1.89 & 18909 & 1.90 & 0.053 & 5409 & 0.055\\
820 & 100 & 14 & 84124 & 707508737 & 17590458 & 0 & 147 & 5 & 1.89 & 18900 & 1.89 & 0.053 & 5406 & 0.055\\
830 & 100 & 14 & 59627 & 555090629 & 10432224 & 0 & 144 & 4 & 1.88 & 18889 & 1.89 & 0.053 & 5406 & 0.055\\
840 & 100 & 7 & 63708 & 445321606 & 9724008 & 0 & 150 & 5 & 1.88 & 18800 & 1.88 & 0.053 & 5412 & 0.055\\
850 & 100 & 1 & 2161557 & 411722431 & 9373523 & 0 & 126 & 4 & 1.88 & 18800 & 1.88 & 0.053 & 5440 & 0.055\\
860 & 100 & 0 & 1712546 & 318509553 & 6217803 & 0 & 147 & 4 & 1.87 & 18792 & 1.88 & 0.054 & 5489 & 0.056\\
870 & 100 & 2 & 490373 & 230066948 & 4190078 & 0 & 161 & 6 & 1.87 & 18703 & 1.88 & 0.054 & 5594 & 0.057\\
880 & 100 & 2 & 238364 & 157742498 & 2961829 & 0 & 133 & 4 & 1.87 & 18700 & 1.87 & 0.055 & 5611 & 0.057\\
890 & 100 & 0 & 856509 & 128046164 & 2123934 & 0 & 155 & 5 & 1.87 & 18700 & 1.87 & 0.055 & 5634 & 0.058\\
900 & 100 & 0 & 780008 & 108286279 & 1347419 & 0 & 151 & 4 & 1.86 & 18668 & 1.87 & 0.055 & 5653 & 0.058\\
910 & 100 & 0 & 760378 & 97310239 & 1244654 & 0 & 130 & 4 & 1.86 & 18600 & 1.86 & 0.055 & 5669 & 0.058\\
920 & 100 & 0 & 630861 & 87290546 & 1116245 & 0 & 173 & 6 & 1.86 & 18600 & 1.86 & 0.056 & 5719 & 0.059\\
930 & 100 & 0 & 554824 & 76720455 & 1029804 & 0 & 168 & 5 & 1.86 & 18600 & 1.86 & 0.057 & 5832 & 0.060\\
940 & 100 & 0 & 496870 & 66662960 & 868363 & 0 & 186 & 6 & 1.86 & 18600 & 1.86 & 0.058 & 5916 & 0.061\\
950 & 100 & 0 & 459163 & 59376307 & 777377 & 0 & 181 & 5 & 1.85 & 18506 & 1.86 & 0.058 & 5962 & 0.061\\
960 & 100 & 0 & 440363 & 53499751 & 727897 & 0 & 177 & 6 & 1.85 & 18500 & 1.85 & 0.059 & 5981 & 0.061\\
970 & 100 & 0 & 382836 & 49044965 & 655341 & 0 & 144 & 5 & 1.85 & 18500 & 1.85 & 0.059 & 6023 & 0.062\\
980 & 100 & 0 & 380332 & 45898581 & 574910 & 0 & 172 & 6 & 1.85 & 18500 & 1.85 & 0.060 & 6104 & 0.062\\
990 & 100 & 0 & 344509 & 41741061 & 537047 & 0 & 175 & 5 & 1.85 & 18500 & 1.85 & 0.061 & 6200 & 0.063\\
1000 & 100 & 0 & 325419 & 38779506 & 466097 & 0 & 235 & 7 & 1.84 & 18485 & 1.85 & 0.062 & 6358 & 0.065\\
\hline
\end{tabular}
\end{table}

\newpage
\begin{table}
\centering
\caption{Full results of the No-Triangle SAT algorithm for N=225.}
\begin{tabular}{|r|r|r|r|r|r|r|r|r|r|r|r|r|r|r|}
\hline
675 & 25 & 25 & 68 & 10099 & 2231 & 0 & 30 & 3 & 2.10 & 5252 & 2.11 & 0.060 & 1504 & 0.061\\
685 & 25 & 25 & 86 & 9247 & 1101 & 0 & 24 & 3 & 2.08 & 5224 & 2.09 & 0.058 & 1473 & 0.059\\
695 & 25 & 25 & 74 & 13100 & 2818 & 0 & 39 & 3 & 2.07 & 5197 & 2.08 & 0.057 & 1448 & 0.058\\
705 & 25 & 25 & 57 & 17163 & 2175 & 0 & 34 & 5 & 2.06 & 5173 & 2.07 & 0.057 & 1427 & 0.058\\
715 & 25 & 25 & 67 & 22706 & 3682 & 0 & 30 & 4 & 2.05 & 5149 & 2.06 & 0.056 & 1406 & 0.057\\
725 & 25 & 25 & 94 & 41106 & 6085 & 0 & 37 & 4 & 2.04 & 5123 & 2.05 & 0.055 & 1389 & 0.056\\
735 & 25 & 25 & 545 & 77983 & 12472 & 0 & 26 & 3 & 2.03 & 5096 & 2.04 & 0.054 & 1375 & 0.056\\
745 & 25 & 25 & 226 & 70447 & 12598 & 0 & 33 & 4 & 2.02 & 5073 & 2.03 & 0.054 & 1359 & 0.055\\
755 & 25 & 25 & 1005 & 156997 & 21906 & 0 & 39 & 4 & 2.02 & 5050 & 2.02 & 0.054 & 1358 & 0.055\\
765 & 25 & 25 & 664 & 194346 & 27684 & 0 & 51 & 6 & 2.01 & 5026 & 2.02 & 0.053 & 1341 & 0.054\\
775 & 25 & 25 & 325 & 399061 & 182180 & 0 & 40 & 4 & 2.00 & 5004 & 2.01 & 0.052 & 1322 & 0.054\\
785 & 25 & 25 & 516 & 526834 & 104346 & 0 & 32 & 3 & 1.99 & 4986 & 2.00 & 0.051 & 1305 & 0.053\\
795 & 25 & 25 & 367 & 811282 & 179971 & 0 & 41 & 3 & 1.98 & 4973 & 1.99 & 0.051 & 1291 & 0.052\\
805 & 25 & 25 & 1948 & 889473 & 129537 & 0 & 34 & 4 & 1.98 & 4950 & 1.98 & 0.051 & 1282 & 0.052\\
815 & 25 & 25 & 1830 & 2743849 & 511557 & 0 & 42 & 3 & 1.97 & 4930 & 1.98 & 0.050 & 1279 & 0.052\\
825 & 25 & 25 & 1012 & 4418564 & 668641 & 0 & 43 & 4 & 1.96 & 4923 & 1.97 & 0.051 & 1289 & 0.052\\
835 & 25 & 25 & 6416 & 7745404 & 1182193 & 0 & 57 & 5 & 1.96 & 4900 & 1.96 & 0.050 & 1277 & 0.052\\
845 & 25 & 25 & 251756 & 20618063 & 3119407 & 0 & 35 & 4 & 1.95 & 4887 & 1.96 & 0.050 & 1266 & 0.051\\
855 & 25 & 25 & 28762 & 42204876 & 14326970 & 0 & 45 & 7 & 1.95 & 4875 & 1.95 & 0.050 & 1259 & 0.051\\
865 & 25 & 25 & 7148 & 65319255 & 18582405 & 0 & 42 & 4 & 1.94 & 4857 & 1.95 & 0.049 & 1248 & 0.051\\
875 & 25 & 22 & 156150 & 306679993 & 63315167 & 0 & 38 & 4 & 1.94 & 4850 & 1.94 & 0.049 & 1247 & 0.051\\
885 & 25 & 13 & 1591849 & 788038314 & 60397295 & 0 & 42 & 4 & 1.93 & 4830 & 1.94 & 0.049 & 1243 & 0.050\\
895 & 25 & 11 & 17476 & 772210331 & 52629470 & 0 & 46 & 5 & 1.93 & 4825 & 1.93 & 0.049 & 1244 & 0.050\\
905 & 25 & 3 & 1343366 & 858005875 & 44622644 & 0 & 54 & 5 & 1.92 & 4812 & 1.93 & 0.050 & 1258 & 0.051\\
915 & 25 & 4 & 604105 & 718516087 & 38708440 & 0 & 54 & 6 & 1.92 & 4800 & 1.92 & 0.049 & 1248 & 0.051\\
925 & 25 & 0 & 21929255 & 748475514 & 34726145 & 0 & 62 & 9 & 1.91 & 4799 & 1.92 & 0.049 & 1247 & 0.051\\
935 & 25 & 0 & 24184043 & 704517010 & 31124675 & 0 & 49 & 4 & 1.91 & 4777 & 1.92 & 0.049 & 1241 & 0.050\\
945 & 25 & 0 & 22659603 & 638177233 & 30156699 & 0 & 56 & 4 & 1.91 & 4775 & 1.91 & 0.048 & 1236 & 0.050\\
955 & 25 & 0 & 16249420 & 609166985 & 28810905 & 0 & 50 & 7 & 1.90 & 4769 & 1.91 & 0.049 & 1240 & 0.051\\
965 & 25 & 1 & 4962911 & 511347497 & 24479783 & 0 & 46 & 4 & 1.90 & 4750 & 1.90 & 0.049 & 1246 & 0.051\\
975 & 25 & 0 & 12643543 & 507406618 & 24573862 & 0 & 54 & 6 & 1.90 & 4750 & 1.90 & 0.050 & 1267 & 0.051\\
985 & 25 & 0 & 10500301 & 410234852 & 21394493 & 0 & 54 & 6 & 1.89 & 4748 & 1.90 & 0.050 & 1267 & 0.051\\
995 & 25 & 0 & 5730065 & 318041287 & 18685310 & 0 & 70 & 5 & 1.89 & 4726 & 1.90 & 0.050 & 1265 & 0.051\\
1005 & 25 & 0 & 6884868 & 283880075 & 15595774 & 0 & 52 & 4 & 1.89 & 4725 & 1.89 & 0.050 & 1265 & 0.051\\
1015 & 25 & 0 & 6235113 & 250031251 & 14612456 & 0 & 65 & 6 & 1.89 & 4725 & 1.89 & 0.050 & 1267 & 0.051\\
1025 & 25 & 0 & 5019300 & 196195331 & 11284038 & 0 & 70 & 7 & 1.88 & 4715 & 1.89 & 0.050 & 1267 & 0.052\\
1035 & 25 & 0 & 2866480 & 163642207 & 9445875 & 0 & 67 & 5 & 1.88 & 4700 & 1.88 & 0.050 & 1270 & 0.052\\
1045 & 25 & 0 & 2581415 & 126487598 & 7529615 & 0 & 66 & 6 & 1.88 & 4700 & 1.88 & 0.051 & 1288 & 0.052\\
1055 & 25 & 0 & 2578222 & 100366781 & 6906600 & 0 & 85 & 8 & 1.88 & 4700 & 1.88 & 0.051 & 1308 & 0.053\\
1065 & 25 & 0 & 2043869 & 73964238 & 4423565 & 0 & 73 & 6 & 1.88 & 4700 & 1.88 & 0.052 & 1313 & 0.053\\
1075 & 25 & 0 & 1348411 & 60677064 & 3895537 & 0 & 66 & 5 & 1.87 & 4675 & 1.87 & 0.052 & 1314 & 0.054\\
1085 & 25 & 0 & 1342830 & 55097337 & 5192919 & 0 & 75 & 7 & 1.87 & 4675 & 1.87 & 0.051 & 1320 & 0.054\\
1095 & 25 & 0 & 1120126 & 43772330 & 3239436 & 0 & 64 & 6 & 1.87 & 4675 & 1.87 & 0.052 & 1324 & 0.054\\
1105 & 25 & 0 & 979929 & 39216020 & 2521350 & 0 & 76 & 7 & 1.87 & 4675 & 1.87 & 0.052 & 1326 & 0.054\\
1115 & 25 & 0 & 946247 & 32348483 & 1819514 & 0 & 70 & 6 & 1.87 & 4675 & 1.87 & 0.053 & 1343 & 0.054\\
1125 & 25 & 0 & 803117 & 26630337 & 1349768 & 0 & 94 & 7 & 1.86 & 4657 & 1.87 & 0.054 & 1375 & 0.056\\
\hline
\end{tabular}
\end{table}

\end{document}